\documentclass{article}
\usepackage[utf8]{inputenc}
\usepackage{authblk}
\usepackage{float}
\usepackage{xcolor}
\usepackage{tikz}
\usepackage{soul}
\usepackage{smartdiagram}
\usepackage{caption}
\usepackage{arev}
\usetikzlibrary{mindmap}
\usepackage{pdflscape}
\usepackage{hyperref}
\usepackage{subcaption}
\usepackage{graphicx}
\usepackage{rotating}
\usepackage{comment}
\usepackage{multirow}
\usepackage{lineno}
\usepackage{array}
\usepackage{xcolor}
\usepackage{lscape}
\usepackage[sorting=none]{biblatex}
\usepackage{setspace}
\DeclareUnicodeCharacter{0301}{\'{o}}
\usepackage{amssymb}
\usepackage[a4paper, margin=1in]{geometry}
\usepackage{times}

\makeatletter
\renewcommand{\maketitle}{\bgroup\setlength{\parindent}{0pt}
\begin{flushleft}
  \textbf{\@title}

  \@author
\end{flushleft}\egroup
}
\makeatother

\providecommand{\keywords}[1]
{
  \textbf{\textit{Keywords---}} #1
}

\title{Common Steps in Machine Learning Might Hinder The Explainability Aims in Medicine}
\date{}
\author[1,2,3,4]{Ahmed M Salih}
\affil[1]{Department of Population Health Sciences, University of Leicester, University Rd, LE1 7RH, Leicester, UK}
\affil[2]{William Harvey Research Institute, NIHR Barts Biomedical Research Centre, Queen Mary University of London, Charterhouse Square, London, EC1M 6BQ, London, UK}
\affil[3]{Barts Heart Centre, St Bartholomew’s Hospital, Barts Health NHS Trust, West Smithfield, London, EC1A 7BE, UK}
\affil[4]{Department of Computer Science, University of Zakho, Duhok road, Zakho, Kurdistan, Iraq}

\begin{document}
\maketitle
\thispagestyle{empty} 

\noindent

\begin{abstract}
\noindent
Data pre-processing is a significant step in machine learning to improve the performance of the model and decreases the running time. This might include dealing with missing values, outliers detection and removing, data augmentation, dimensionality reduction, data normalization and handling the impact of confounding variables. Although it is found the steps improve the accuracy of the model, but they might hinder the explainability of the model if they are not carefully considered especially in medicine. They might block new findings when missing values and outliers removal are implemented inappropriately. In addition, they might make the model unfair against all the groups in the model when making the decision. Moreover, they turn the features into unitless and clinically meaningless and consequently not explainable. This paper discusses the common steps of the data pre-processing in machine learning and their impacts on the explainability and interpretability of the model. Finally, the paper discusses some possible solutions that improve the performance of the model while not decreasing its explainability.
\end{abstract}
\newpage

\section{Introduction}
Data pre-processing is an initial indispensable step in machine learning and data science. It consists of several components and processing steps that are performed on raw data to ensure its quality before fitting them to any model. It might involves dealing with missing values, outliers detection and removal, normalization and standardization, dimensionality reduction (e.g., feature selection, principal component analysis), data augmentation of imbalanced data, and dealing with confounding variables~\cite{kang2018machine}. All these steps and more are conducted to ensure the quality of the data which might improve the performance of the model and decrease the running time.\\
On the other hand, explainable artificial intelligence (XAI) as an emerging topic aims to understand how the model works. Its direct aims are more to do with improving the interpretability of the model than improving the performance of the model. It involves quantifying the uncertainty of a machine learning model when making a decision. Moreover, it helps to identify and highlight the most informative features (pixels in an image) in the model that drive its outcome~\cite{salih2024review}. Furthermore, it aims to improve model fairness toward all groups (e.g., sex, ethnicity, unemployment, etc) in the model. The ultimate aim of the XAI is to make the model more transparent and consequently trustfully.\\
Data pre-processing and XAI should not hinder each other. They should rather work together to improve the performance of the model and its explainability simultaneously. In other words, any step or steps to improve the performance of the model should not affect its explainability negatively. However, data-preprocessing might hinder the aims of the XAI and eventually its interpretability. This include how to deal with missing values and which method should be considered for imputation. Moreover, outliers should not be ignored and removed from the model in the medicine domain as they might represent a new case or at least should be explained why they are with extreme values. In addition, dimensionality reduction methods might lead to remove clinical significant features or transfer them into unitless which make them difficult to interpret. Furthermore, the data normalization and standardization might improve the running time and model performance, but they turn the features into clinically meaningless. Data augmentation should be applied appropriately and the augmentation should consider the structure of the data if they came from different groups (e.g., sex, ethnicity). Finally, how to deal with the group of confounds as the adopted method might have positive/negative impact on XAI.\\
This paper discusses some essential steps of data pre-processing in machine learning and how they might improve the accuracy of the model but hinder its explainability.
\section{Data preprocessing steps}
Most of the data cannot be used directly in the model and there are some essential data pre-processing steps need to be conducted before fitting the data to the model. The steps might be different based on the type of the data, number of data modality, sample size and the used model. Below we discuss the most common steps of data preprocessing which they are more suitable with tabular data. The steps include missing values, outliers removal, data augmentation, normalization and standardization, dimensionality reduction methods (feature selection and principal component analysis) and confounding variables.
\subsection{Missing values}
\begin{figure}[H]
\centering
    \includegraphics[width=0.4\linewidth]{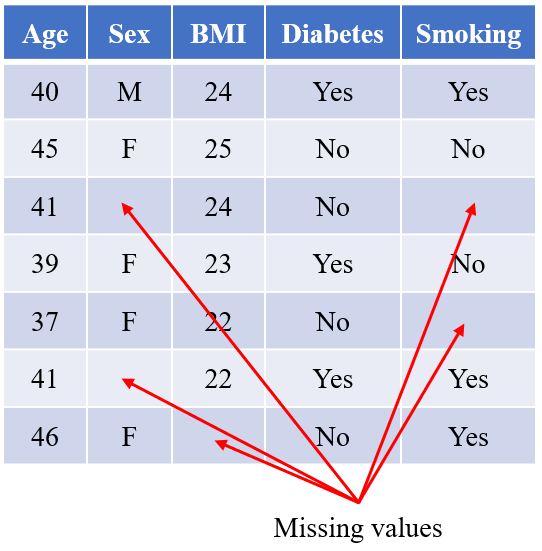}
    \caption{Several missing values in tabular data.}
    \label{Missingv}
\end{figure}

Missing values is one the most prevalent issue in machine learning especially in medicine which might impair the explainability of the model. Usually the data are not complete for all individuals for variety of reasons. Possible way to deal with the missing values is to remove them either at individuals level or at features level when the missing rate is high based on a pre-determined threshold. Other common way is to impute those missing values using either the mode, median or mean value based on the data type and the distribution of the data. Figure~\ref{Missingv} shows the missing values for some individuals for some features.
\noindent
XAI and interpretability is significantly affected by the chosen method to impute the missing values~\cite{vo2024explainability}. Moreover, as the real value is unknown, using those imputation methods might lead to a case which is called counterfactuals~\cite{ahmad2019challenge}. Two studies~\cite{ahmad2019challenge, vo2024explainability} discussed how the imputation method might alter the explanation and consequently might have negative impact on the human well being and safety. In addition, the negative impact of the missing value on the explainability increases when the missing rate increases as well. Accordingly, dealing with the missing values while keeping it explainable might need a careful consideration.
\subsection{Outliers removal}
\begin{figure}[H]
\centering
    \includegraphics[width=0.35\linewidth]{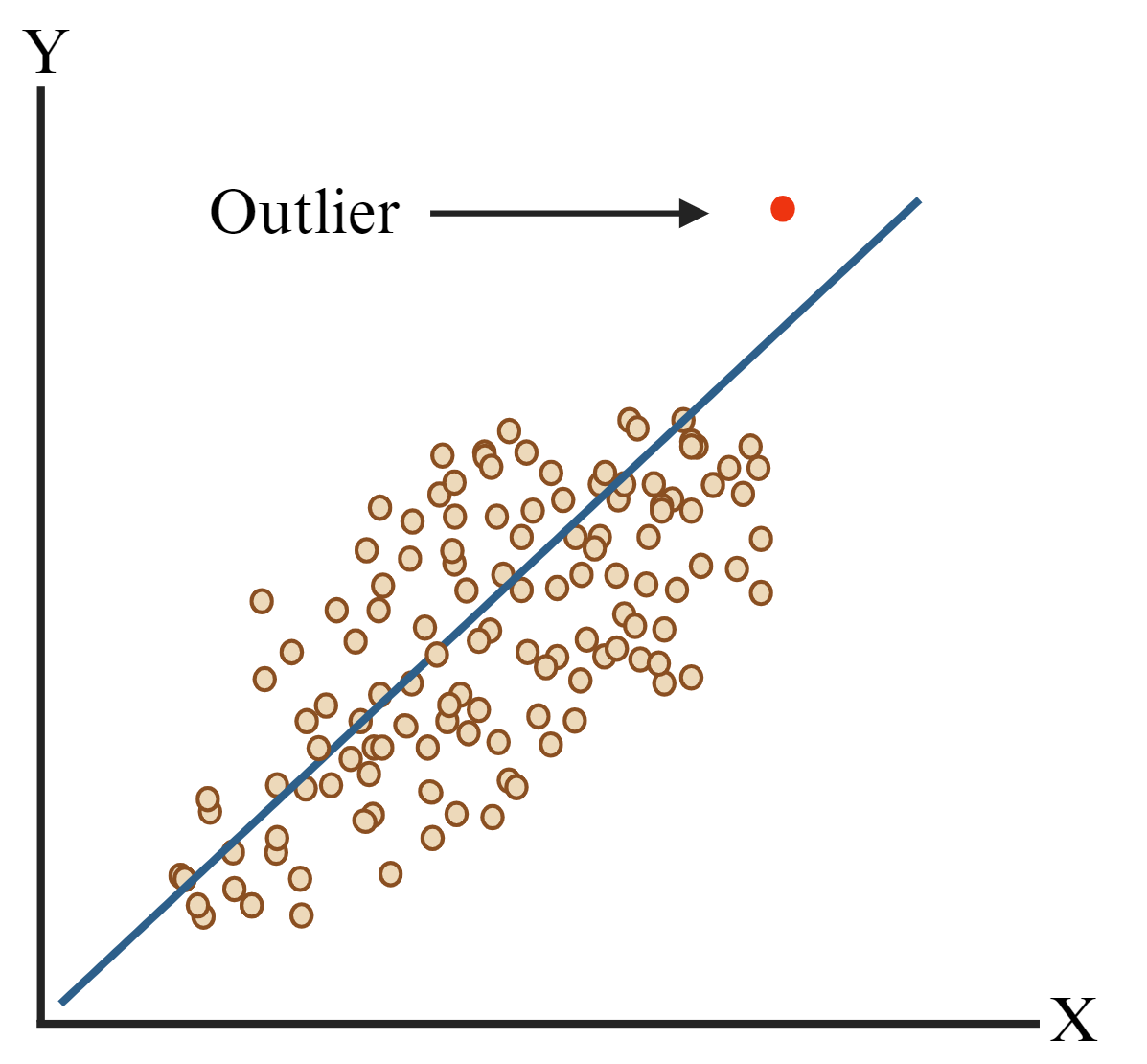}
    \caption{An outlier that is deviated from the population.}
    \label{outlier}
\end{figure}

One of the most common step in machine learning modelling is outliers detection and removal as they might affect negatively the model performance. It means removing data points with extreme large or small values~\cite{smiti2020critical} as it is shown in figure~\ref{outlier}. There are multiple methods to identify the outliers for exclusion including those based on statistical methods, distance-based, density-based and cluster-based~\cite{smiti2020critical}.
\noindent
However, aligning with the aims of the XAI, every data point which represent valuable information of an individual as a patient or control need to be explained and should not be ignored. Clinically speaking, outliers might hold a clinical interpretation which explain why the values are very big or small. They might represent a new case that has not be investigated before and each time were removed because they were considered as outliers. XAI should be used to explain and interpret why the model detected those data points as outliers and why their values are deviated from the norm values. 
\subsection{Data augmentation}
\begin{figure}[H]
\centering
    \includegraphics[width=0.5\linewidth]{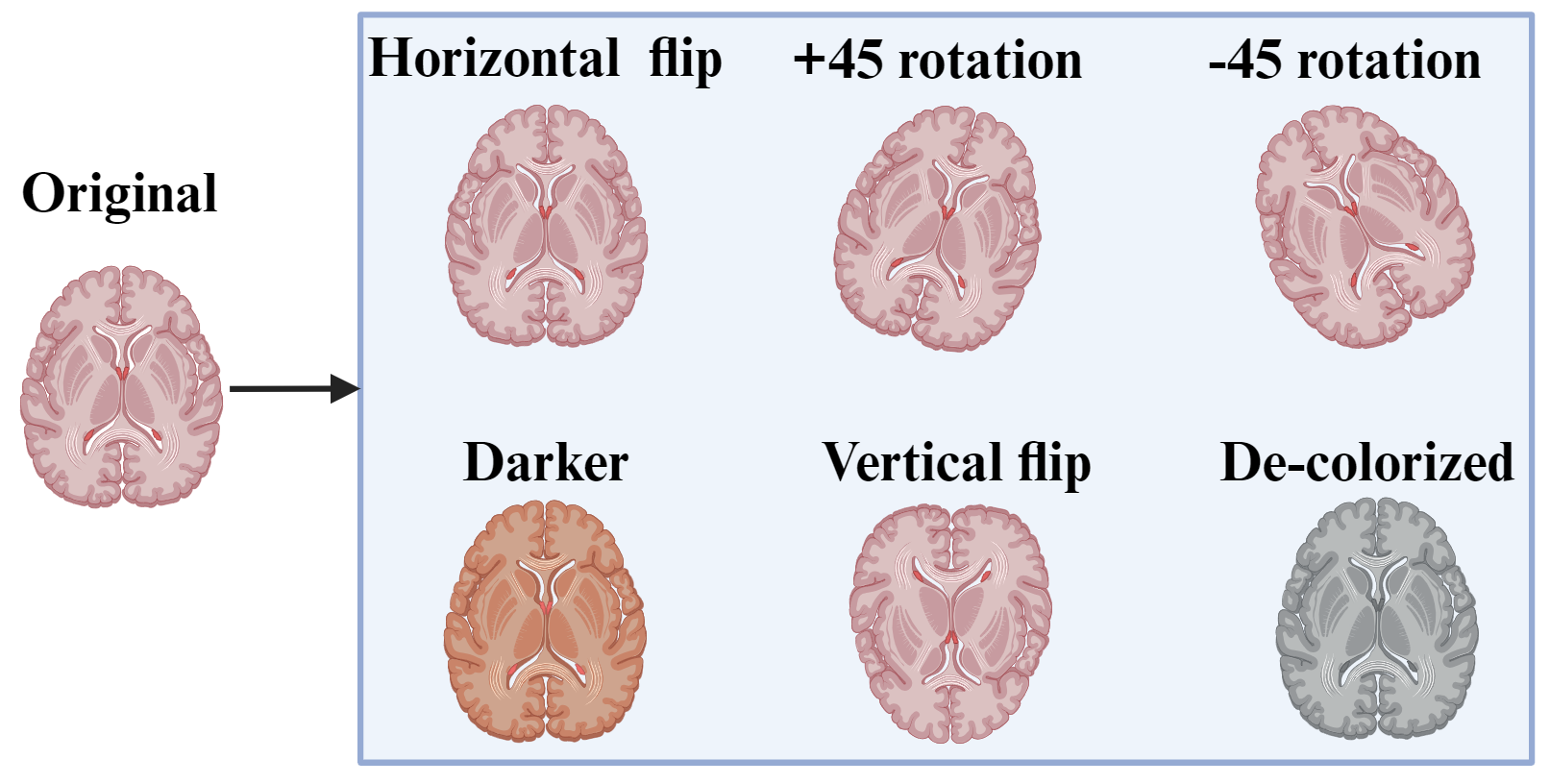}
    \caption{Data augmentation approaches.}
    \label{data_augmen}
\end{figure}

In medicine it is usually the number of cases and control are not similar and they might vary in numbers significantly. This is an issue when applying a classification model which is called imbalanced data. The possible solutions in this case is the data augmentation to increase the number of cases or control to convert them into a more balanced data. This might also include over or under sampling. Figure~\ref{data_augmen} shows a data augmentation where the original image is flipped vertically and horizontally, de-colorized, rotate and make it darker to generate more images from the original one. All these steps were performed to augment the data of the minor class to make it similar to the major class which improve the model performance.
\noindent
Data augmentation should be implemented and interpreted accurately. Considering a case where the number of cases is massively smaller than the number of control for a given population (Alzheimer's disease vs control free of brain diseases). If data augmentation is performed on all data set, then the data set does not represent the population anymore neither the generated explanation from XAI methods can be generalized over the population. On the other hand. if the data augmentation is performed on only the training data, then the test and the training data will have different distribution and the model performance might be poor on the test data. Consequently, different explanation might be obtained when XAI is applied to training and/or test data.\\
The other significant concern of data augmentation for the minor class is how the augmented data represents all the groups in the original data. For instance, assuming the minor class involves data from multiple ethnicity and it is dominated by the White ethnicity. Does the augmented data considered the minor group (e.g., Black, Asian) in the raw data so that the explanation can be generalized for all ethnicities?. Similar concern can be extended to sex and age groups. These points are very critical because they might affect the explanation of the model and provide misleading clinical outcome.
\subsection{Normalization and Standardization}
Normalization might be one of the last common process to perform before fitting the data into the machine learning model. It is the process of transferring the data into a common range scale to avoid the domination of the features with high values over those with small values~\cite{singh2020investigating}. There have been many methods proposed to normalize the data each using different statistical measures. This include mean and standard deviation-based, median-based, mini-max-based, tanh-based, decimal scaling-based and sigmoidal~\cite{singh2020investigating}.\\
Standardization on the other hand is the process of converting the data to have a zero mean and a unit of standard deviation. To standardize the data, the mean is subtracted from each data point and then divided by the standard deviation. Figure~\ref{Missingv} shows a visual representation on how the data might look like when normalization and/or standardization are performed.
\begin{figure}[H]
\centering
    \includegraphics[width=0.5\linewidth]{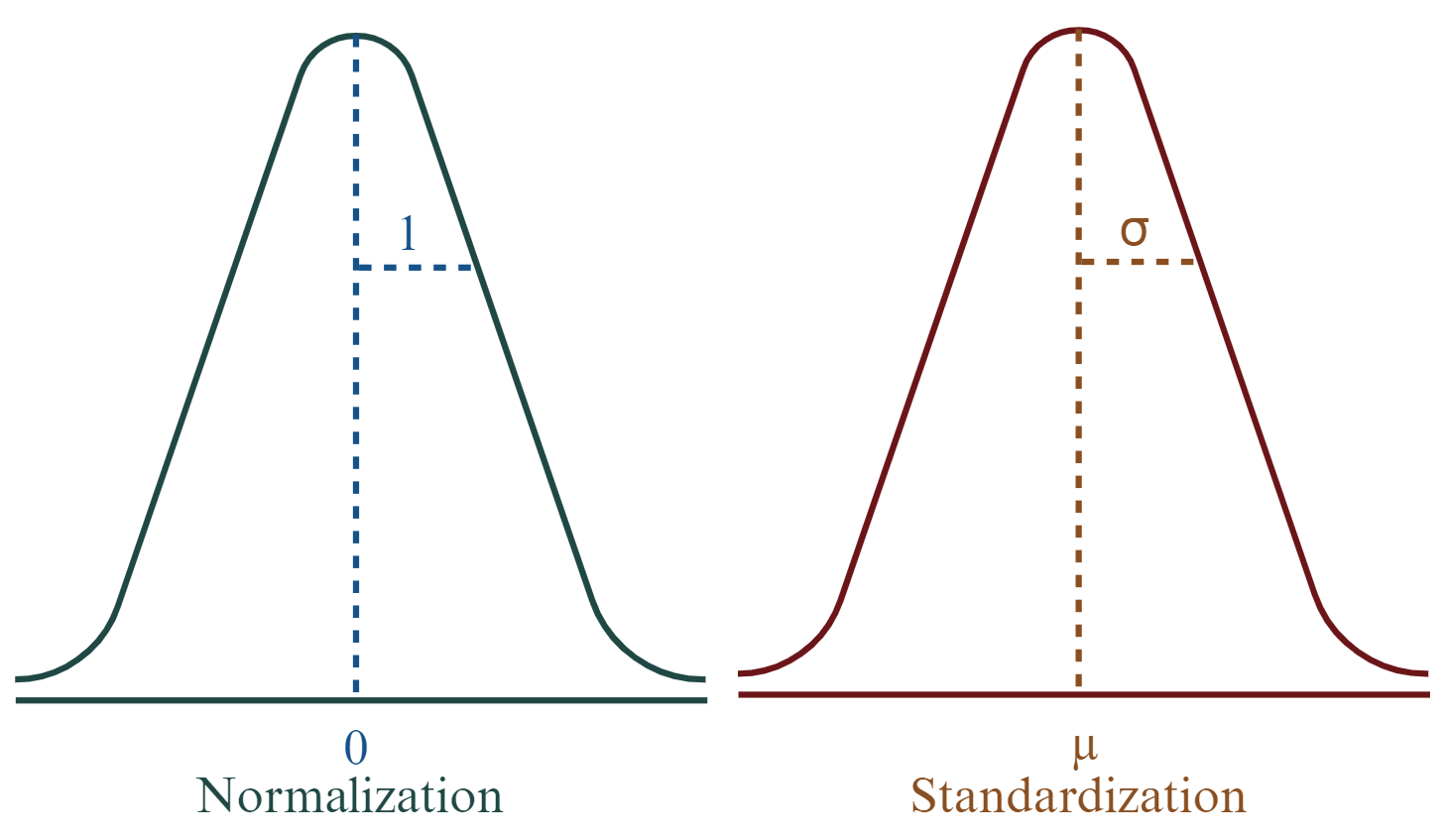}
    \caption{Visual representation of normalization and standardization methods.}
    \label{NOR_vs_STA}
\end{figure}
\noindent
Before emerging XAI and its aims, the accuracy of the model was the most significant aim to achieve and any step that improves the model performance should be conducted. However, with XAI it is important to understand how the model works and reveal the impact of the most informative features. While the two aforementioned pre-processing steps have shown to be effective and increase the model performance, they hinder the explainability and the interpretability of the model. This is because the transformed features are unitless and the effect size of the features cannot be explained in the original unit of the feature~\cite{salih2024linear}. For instance, it is really vital to explain to a patient or a clinician how increasing the weight by one kilogram would increases the risk of a cardiovascular disease. Furthermore, assuming a feature (among others) represents the sleeping duration in minutes or hours in a model with mental well being or dementia as outcome. It is more informative that XAI shows how more/less minutes/hours of sleeping would increases/decrease the risk of the well being/dementia. However, with data normalization such explanation is not possible anymore because the new version of the features are unitless. On the other hand, performing standardization on the raw data might also be a factor of obstacle to explain the model and the features in unite. This is because the change in the standardized data represents the variation from the mean or the center of the data. Accordingly, the effect size of the feature cannot be explained in the original unit for an individual.
\subsection{Feature selection}
\begin{figure}[H]
\centering
    \includegraphics[width=0.6\linewidth]{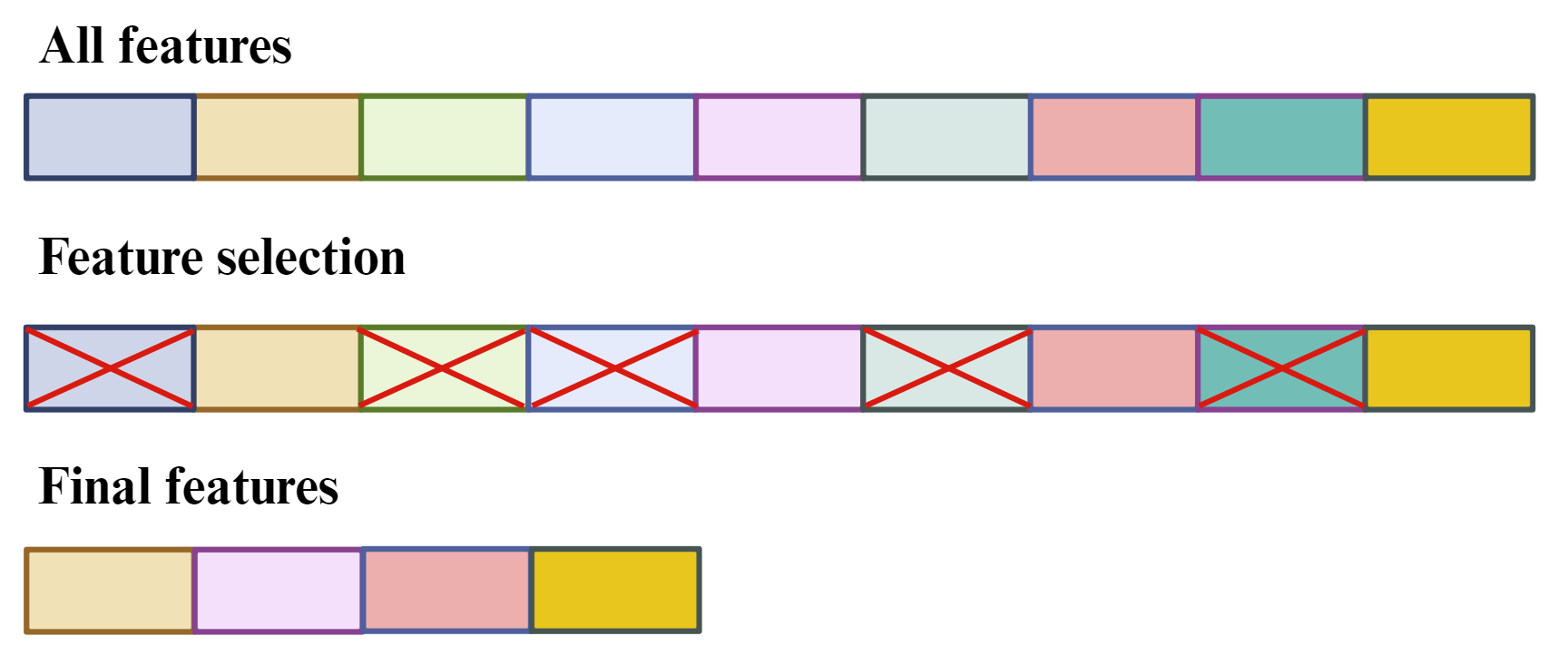}
    \caption{Feature selection steps and the final set.}
    \label{feautre_sele}
\end{figure}

One of the most common step in machine learning modelling is feature selection. It is especially recommended when the number of features is high or the the features are collinear~\cite{cai2018feature}. It reduces the dimension of the features and consequently might decrease the running time. Moreover, it has the ability to remove noisy or irrelevant features which might improve the model performance. Figure~\ref{feautre_sele} shows common steps of feature selection where some of the features might be removed while those left are used in the machine learning model. There are many methods to perform feature selection including those based on filter methods, embedded method, clustering and ensemble method~\cite{chandrashekar2014survey}.
\noindent
Although feature selection is considered one of the main step to remove noisy, redundant and irrelevant features, it might prevent or block some clinical explanation. For instance, glucose and insulin are found to be associated positively in the blood in a non-diabetic individual~\cite{nakrani2020physiology}. Feature selection might pick one of them and consider the other as a redundant features because they are highly correlated. However, the clinical interpretation of each one of them might be different when they appear among the most informative features in a machine learning model (diabetes vs control) after applying an XAI method. For example, non-diabetic hyperglycemia is a condition related to  impaired glucose regulation in individuals without diabetes~\cite{fattorusso2018non}. Other example could the association of ApoE4 with t-tau protein. It has been found that there is a significant association between the two in individuals with Alzheimer’s disease and dementia~\cite{benson2022don}. One of them might be picked by a feature selection method while the other will be removed as it will be considered as a redundant. However, the clinical interpretation of each of the two and their impact toward brain diseases might be different.
\subsection{Principal component analysis}
\begin{figure}[H]
\centering
    \includegraphics[width=0.4\linewidth]{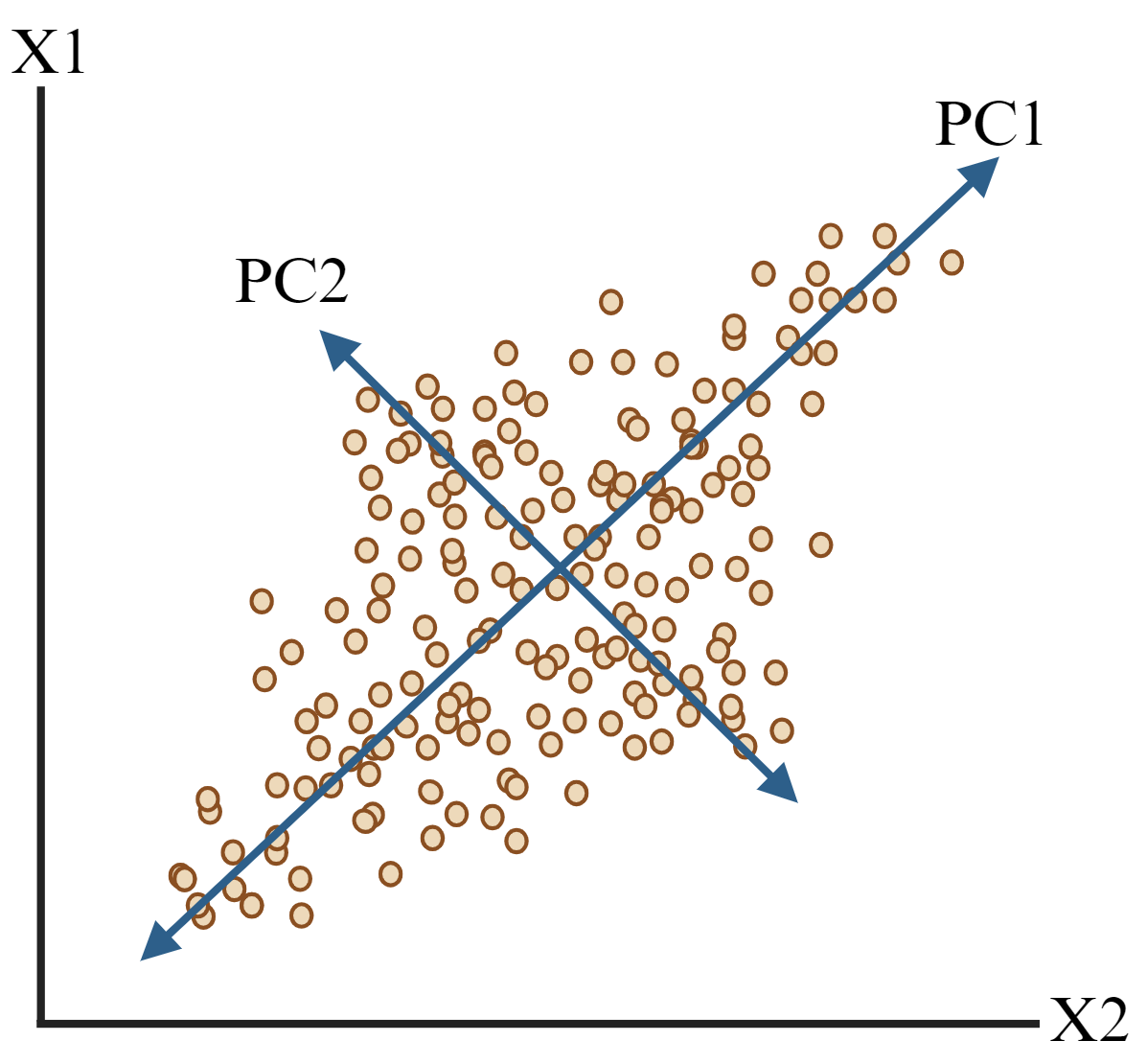}
    \caption{Generates new components from correlated features.}
    \label{PCA}
\end{figure}

Principal component analysis (PCA) is one of the most common method used in machine learning as a dimensionality reduction technique. It transfers the original data to new components that are linearly uncorrelated and capture the most variation in the original data. The number of components can be chosen manually or automatically based on the captured variation in the original space. It has several advantages toward machine learning model performance including reducing running time and removing the impact of collinearity. Figure~\ref{PCA} shows an example of how a component might be generated from two correlated features.\\
\noindent
Although applying PCA might improve the performance of the machine learning model, its implementation hinders achieving the principles of XAI for several reasons. Firstly, applying PCA means removing the unit of the features and the coefficient value of these new components in machine learning model are unitless. For instance, if the number of original features are ten and they have been transferred into five independent components. After applying an XAI method, it is found that the component one is the most informative one toward model decision. It is really not clear how to interpret the outcome of XAI in units because the identified component is generated from several features each with its own contribution and unit. Secondly, it is challenging to even identifying the most informative features in the model after they have been transferred into PCs. Some might argue that the new components might be returned into their original space and accordingly will be able to identify the most informative features based on their contributions in each component. However, the contribution of each feature is allocated into several components and it is challenging to be precise how exactly they contribute in the model outcome.

\subsection{Confounding variables}
\begin{figure}[H]
\centering
    \includegraphics[width=0.7\linewidth]{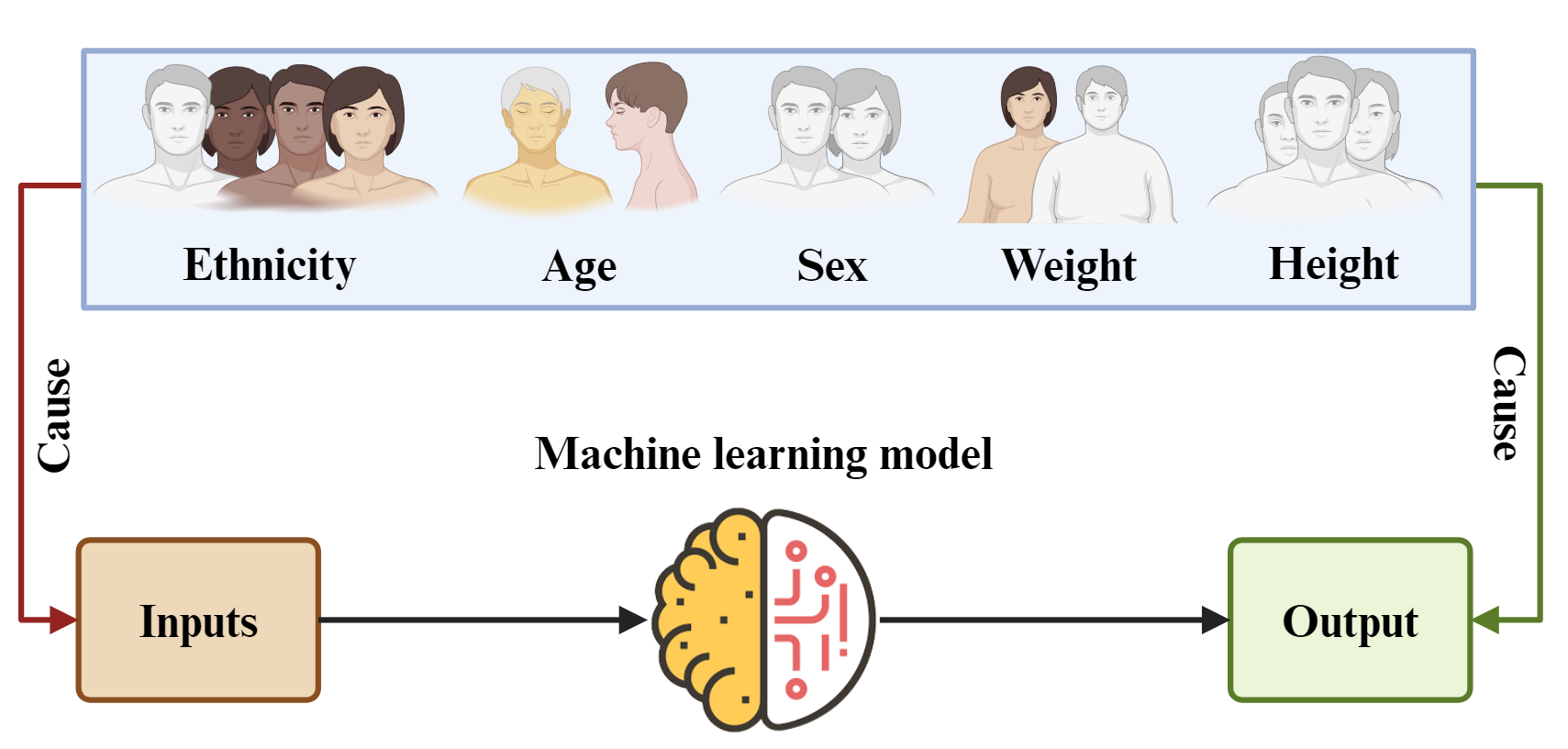}
    \caption{A set of the most common confounding variables in medicine and biomedical studies.}
    \label{Covariates_mv}
\end{figure}

In most of the clinical researches there are confounding variables. They are a group of variables that affect the input data and the output simultaneously. They might be at individual levels such as sex, ethnicity, age, weight and height or they could be related to data acquisition as the case with brain Magnetic resonance imaging~\cite{alfaro2021confound}. Within medicine, the set of the confounding variables is different from a case to another based on the kind of data and the examined condition. However, the confounding variables in the figure~\ref{Covariates_mv} are considered in most of medicine researches unless it is the examined condition.
\noindent
Firstly, it is not an easy task to define the set of the confounding variables, especially in the medicine. It is confusing whether to consider a specific variable as a confounding variable or as an independent variable. Secondly, how dealing with the confounding variables might have significant impact on the explainability and the interpretation of the model. For instance, one of the approaches to mitigate the impact of the confounding variables is by applying a matching method (e.g., propensity score) where cases and control are matched based on sets of confounding variables. Although this method might reduces the impact of the confounding variables, however it might block a clinical clue which might be captured by XAI methods. For example, it is well established of sex differences in the risk of cardiovascular diseases~\cite{humphries2017sex} and neurodegenerative~\cite{vegeto2020role}. Moreover, ethnic minority groups including Asian and Black are more at risk of infection of COVID-19 than White ethnicity~\cite{sze2020ethnicity}. When applying the matching method using confounding variables (ethnicity and sex among them), XAI cannot capture of the ethnicity and sex differences in these two conditions.\\
Another common approach is to regress the confounding variables from the independent variables before fitting them in the machine learning model~\cite{dinga2020controlling}. However such approach does not guarantee the elimination of the effects of the confounding variables and the model outcome might be biased~\cite{dinga2020controlling, spisak2022statistical}.
\section{Discussion and Recommendations}
To conclude, there is no a magic solution and each method or approach to adopt to improve the model explainability has its own strengths and weaknesses. The paper discussed several steps that might improve the performance of the machine learning models, but they might hinder their explainability. The current trends in developing machine learning system particularly in medicine is that they should be explainable and interpretable so both the patients and the clinicians understand the AI decision.

\begin{enumerate}
    \item \textbf{Missing values:} In some medicine cases removing the missing values is not an option because the sample size is small unless the missing rate is extreme. To ensure that an appropriate method is used to impute the missing values, it is vital to use more than one imputation method with several models and XAI to compare and contrast how the explanation will be different. The imputation method that provide a more consistent explanation across the models and XAI should be considered.

    \item \textbf{Outliers removal:}  Considering the aims of XAI in medicine, no data points should be removed and ignored because they represent patients that deserve an explanation why they hold extreme values. Possible solution could be by implementing clustering models where those with extreme values are combined together and modelled separately. Other way could by applying some XAI methods where first they are detected and then explained why they hold such extreme values~\cite{li2023survey}.

    \item \textbf{Data augmentation:} In some scenarios the number of cases and control extremely different especially with rare diseases. This necessitate to implement data augmentation for the minor class. However, the augmentation should consider the structure of the raw data in order to ensure the explanation can be generalized over all minorities. In other words, data augmentation should not decreases the fairness of the model toward the minor groups (e.g., sex, ethnicity, age groups, marital status). Possible solution could be through applying several data augmentation methods and examine how the explanation will be different and compare it with a baseline.

    \item \textbf{Normalization and Standardization:} Although these two steps are used frequently in machine learning modeling, however their impacts on the explanation quality has not be investigated before~\cite{salih2024linear}. Possible current solution might be through adopting machine learning models that are able to deal with the diversity in the range of the data in the original unit without transferring them into unitless. Other possible solution could be through developing some methods that are able to normalize/standardize the data and return them in the original space in order to explain their impacts in the original unit when XAI is implemented.

    \item\textbf{Dimensionality reduction:} Feature selection and PCA are the most two common techniques used to reduce the dimension of the data or remove correlated, noisy and irrelevant features. However, these two methods should not hinder the model explainability and they should be adopted appropriately. Feature selection might be more appropriate and inline with the aims of XAI but the selection should not be based only on the statistical criterion. Clinical information might also be used to select the features and the one has more association with the outcome statistically and clinically should be considered. PCA might be less appropriate in medicine because of the reasons mentioned before when the aim is to make the model more explainable than improving its performance.

    \item\textbf{Confounding variables:} There might not be a golden option to consider in all health applications to mitigate the impacts of the confounding variables and do not loss the explainability of the model. It highly depends on the outcome and the sets of the variables to consider as confounding variables. One of the approaches is to include them in the model alongside the independent variables. While this would help to explain the model when switching between their values (male to female, or between ethnic groups), however the classic interpretation (holding other variables as constant) of the coefficient value in the model makes it difficult to understand how the values of the independent features will be changed.
    
\end{enumerate}
\section{Acknowledgments}
AMS acknowledges support from The Leicester City Football Club (LCFC). All figures are generated by Biorender~(\url{https://www.biorender.com}).

\printbibliography
\end{document}